\pdfoutput=1

\documentclass[11pt]{article}

\usepackage[preprint]{acl}

\usepackage{times}
\usepackage{latexsym}

\usepackage[T1]{fontenc}

\usepackage[utf8]{inputenc}

\usepackage{microtype}

\usepackage{inconsolata}

\usepackage{graphicx}

\usepackage{enumitem}
\usepackage{fancyvrb}
\usepackage{listings}
\usepackage{multirow}
\usepackage{multicol}
\usepackage{booktabs}
\usepackage{amsmath}
\usepackage{amsfonts}

\usepackage[most]{tcolorbox}
\usepackage{xcolor} 

\definecolor{c1}{cmyk}{0,0.6175,0.8848,0.1490} 
\definecolor{c2}{cmyk}{0.1127,0.6690,0,0.4431} 
\definecolor{c3}{cmyk}{0.3081,0,0.7209,0.3255} 
\definecolor{c4}{cmyk}{0.6765,0.2017,0,0.0667} 
\definecolor{c5}{cmyk}{0,0.8765,0.7099,0.3647} 
\definecolor{forestgreen}{HTML}{397727} 

\newtcbox{\hlprimarytab}{on line, rounded corners, box align=base, 
  colback=c3!10, colframe=white, size=fbox, arc=3pt, 
  before upper=\strut, top=-1pt, bottom=-4pt, left=-2pt, right=-2pt, boxrule=0pt}
\newtcbox{\hlsecondarytab}{on line, box align=base, 
  colback=c1!10, colframe=white, size=fbox, arc=3pt, 
  before upper=\strut, top=-2pt, bottom=-4pt, left=-2pt, right=-2pt, boxrule=0pt}

\newcommand{\uag}[1]{{\hlprimarytab{#1}}}  
\newcommand{\dab}[1]{{\hlsecondarytab{#1}}}  

\title{Pre-Training Curriculum for Multi-Token Prediction in Language Models}

\author{Ansar Aynetdinov \\
  Humboldt-Universität zu Berlin \\
  \texttt{aynetdia@hu-berlin.de} \\\And
  Alan Akbik \\
  Humboldt-Universität zu Berlin \\
  \texttt{alan.akbik@hu-berlin.de} \\}

\begin{document}
\maketitle

\begin{abstract}
Multi-token prediction (MTP) is a recently proposed pre-training objective for language models. Rather than predicting only the next token (NTP), MTP predicts the next \textit{k} tokens at each prediction step, using multiple prediction heads. MTP has shown promise in improving downstream performance, inference speed, and training efficiency, particularly for large models. However, prior work has shown that smaller language models (SLMs) struggle with the MTP objective. To address this, we propose a curriculum learning strategy for MTP training, exploring two variants: a forward curriculum, which gradually increases the complexity of the pre-training objective from NTP to MTP, and a reverse curriculum, which does the opposite. Our experiments show that the forward curriculum enables SLMs to better leverage the MTP objective during pre-training, improving downstream NTP performance and generative output quality, while retaining the benefits of self-speculative decoding. The reverse curriculum achieves stronger NTP performance and output quality, but fails to provide any self-speculative decoding benefits. 

\end{abstract}

\section{Introduction}

In recent years, large language models (LLMs) have demonstrated remarkable capabilities in understanding and generating complex text, code, and other modalities. These advances have been driven primarily by improvements in model architectures, training data quality and scale, and optimization strategies, and yet the underlying objective for most widely-adopted LLMs remains the next-token prediction loss. Models such as GPT \cite{gpt3, gpt4}, LLaMA \cite{llama1, llama2, llama3}, and Phi \cite{phi15, phi3} are conventionally trained to predict a single token at each generation step, implicitly treating language modeling as a sequence of one-step-ahead predictions. While this formulation is simple and effective, it may not fully leverage the predictive capabilities of LLMs or reflect the underlying structure of natural language.

\citet{gloeckle2024mtp} investigated the potential of a multi-token prediction (MTP) objective for LLMs. Instead of predicting just the next token (NTP), their approach aims to produce multiple subsequent tokens at each prediction step, with multiple output heads that are independent from each other but share a common model backbone. They found this approach to improve model's downstream performance, inference speed, and training sample efficiency without significantly increasing training time. Recently, DeepSeek-AI adopted the MTP training objective for their V3 model \cite{deepseek_v3} that serves as a base model for the reasoning R1 model \cite{deepseek_r1}. 

So far, the previous work prioritized mid- and large-sized models with at least 7B parameters, since \citet{gloeckle2024mtp} observed that MTP leads to more performance gains as the model size increases. We hypothesize that smaller language models (SLMs) enjoy less benefits from the MTP objective during pre-training due to the fact that they struggle to deal with morphological and semantic dependencies between multiple tokens at once from the get-go. 

\noindent
\textbf{A curriculum-based approach to MTP.}
Inspired by the work of \citet{bengio2009curriculum}, who were the first to propose curriculum learning strategies in the context of machine learning, we explore using a pre-training curriculum to enable SLMs to better leverage the benefits of the MTP objective. Since the complexity of the MTP objective increases with the amount of tokens considered, it is easy to construct a curriculum by gradually changing the number of tokens considered in the objective.

\begin{figure*}[ht]
  \includegraphics[width=\linewidth]{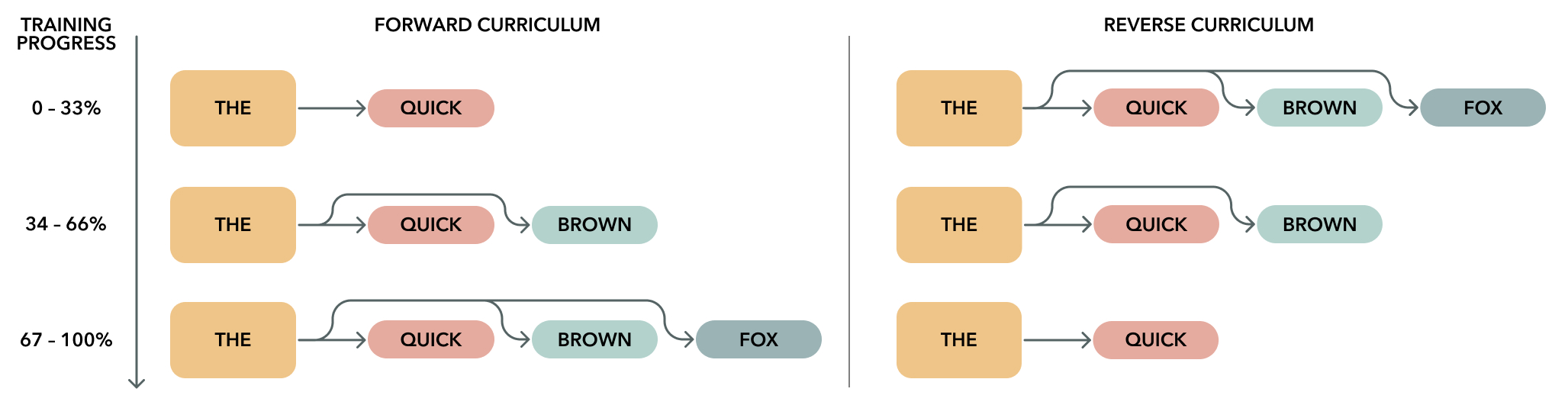}
  \caption{Visualization of the forward and reverse MTP curricula. When training a language model on a 3-token-prediction task for $n$ steps, the forward curriculum starts with a vanilla NTP task, adding an additional token to the task every $\frac{n}{3}$ steps. The reverse curriculum does the opposite, starting with a full 3-token-prediction task, and dropping a token from the task every $\frac{n}{3}$ steps.}
  \label{fig:viz}
\end{figure*}

Figure \ref{fig:viz} illustrates our proposed curriculum variants. Under a forward curriculum, models are given a simpler NTP task in the beginning of pre-training, and as the training progresses, they are guided towards a more complex MTP task. The task is expanded by incrementally including one additional token at uniform intervals throughout pre-training. The reverse curriculum, on the other hand, starts with a full MTP task, and incrementally removes one token from the task also at uniform intervals. The two opposing curricula are designed to clearly determine whether MTP SLMs perform better with an easy-to-hard or a hard-to-easy learning strategy.

\noindent \textbf{Contributions.} This work makes the following contributions:

\begin{itemize}[leftmargin=\parindent]
    \item We provide a comprehensive exploration of the MTP objective's potential when used to train SLMs predominantly on natural language.
    \item We propose and evaluate the validity of novel curriculum-based MTP training strategies.
    \item We showcase that curriculum-based (dynamic) MTP pre-training strategies allow SLMs to learn the MTP task better than static MTP baselines.
\end{itemize}

\section{Preliminaries}

In this section, we formalize the multi-token prediction objective and describe the design of the proposed curricula.

\subsection{Multi-Token Prediction Objective}

Traditional large language models are trained using the next-token prediction (NTP) objective, where the model, given a context sequence \( \mathbf{x} = (x_1, x_2, \dots, x_t) \), is tasked with predicting the next token \( x_{t+1} \). This is accomplished by maximizing the log-likelihood of the target token:
\begin{equation*}
\mathcal{L}_{\text{NTP}} = - \sum_{t=1}^T \log P(x_{t+1} | x_1, \dots, x_t; \theta),
\end{equation*}
where \( \theta \) represents the model parameters.

In contrast, the multi-token prediction (MTP) objective extends this task by requiring the model to predict a sequence of \( k \) tokens \( \mathbf{y} = (x_{t+1}, x_{t+2}, \dots, x_{t+k}) \) simultaneously. The MTP loss is defined as:

\begin{equation*}
\mathcal{L}_{\text{MTP}} = - \sum_{t=1}^T \sum_{i=1}^k \log P(x_{t+i} | x_1, \dots, x_t; \theta),
\end{equation*}

where \( P(x_{t+i} | x_1, \dots, x_t; \theta) \) is computed using multiple output, or language modeling, heads conditioned on the shared model backbone.

Predicting multiple tokens at each prediction step allows for self-speculative decoding \cite{blockwise_parallel_decoding} during inference without the need for an auxiliary model. This can speed up the inference by up to \( k \) times by reducing the number of forward passes needed to generate a sequence. In addition, the MTP objective promotes the learning of richer contextual dependencies, as the model must optimize for multiple interrelated predictions at each step.

\subsection{Forward and Reverse Multi-Token Curricula}

The design of the curricula dynamically adjusts the number of output, or language modeling, heads $k \in \{1, 2, \dots, k_{\max}\}, \quad k \in \mathbb{Z}$, which corresponds to the number of tokens predicted by the model at each prediction step. These adjustments are based on the training epoch $e$, ensuring systematic changes every $E / k_{\max} $ epochs, where $ E $ is the total number of training epochs. While it is certainly possible to come up with a more adaptive curriculum that relies on the training dynamics, we aim to isolate and demonstrate the core benefit of using a curriculum-based MTP objective over a static one. Therefore in this paper we decide to focus on a simple pre-defined curriculum design.

\subsubsection{Forward Multi-Token Curriculum}

The forward multi-token curriculum starts with the simplest prediction task ($k = 1$) and progressively increases the number of prediction heads. Formally, the number of active output heads at training epoch $e$ can be defined as:

\begin{equation*}
    k_{\text{current}}(e) = \min\left(k_{\max}, \left\lfloor \frac{e}{E / k_{\max}} \right\rfloor + 1\right).
\end{equation*}

This gradual increase in complexity ensures that the model first learns fundamental token-by-token predictions before transitioning to more complex multi-token prediction tasks.

\subsubsection{Reverse Multi-Token Curriculum}

In contrast, the reverse multi-token curriculum starts with the maximum number of prediction heads $( k = k_{\max} )$ and gradually reduces the number of active heads over time. The number of active prediction heads at training epoch $ e $ is defined as:

\begin{equation*}
    k_{\text{current}}(e) = \max\left(1, k_{\max} - \left\lfloor \frac{e}{E / k_{\max}} \right\rfloor\right).
\end{equation*}

This strategy relies on the insight reported by \citet{gloeckle2024mtp}, that pre-training with a multi-token objective improves performance on single-token downstream tasks. We hypothesize that by gradually nudging the model towards NTP during the pre-training stage, we will be able to obtain a better main language modeling head. 

\section{Experiments}

In this section, we aim to address the following 3 aspects of our proposed curricula:

\begin{itemize}[leftmargin=\parindent]
    \item \textbf{Main LM head performance.} Since in both the forward and reverse curricula the MTP objective aligns with the NTP objective at some point during pre-training, we examine how the resulting models perform on standard NTP benchmarks using a single (main) language modeling (LM) head.
    \item \textbf{Inference Speed.} One of the main advantages of the ability to output multiple tokens in one prediction step is the fact that it can be used to significantly speed-up the inference by performing self-speculative decoding. Therefore we measure whether the proposed curricula can offer additional speed-ups during inference. 
    \item \textbf{Output Quality.} Lastly, we evaluate generative capabilities of the models trained under our proposed curricula, i.e., the quality of the sequences that they generate. In the end, this is the most important aspect of language models, and therefore should be the main indicator of how useful the proposed curricula are in the context of multi-token prediction.
\end{itemize}

\subsection{Experimental Setup}

\subsubsection{Models}
We conduct our experiments using transformer-based decoder-only LLMs, specifically the configuration used in the Llama model family \cite{llama2}. Regarding the model size, token types, and language modeling heads enabling the MTP, we follow the insights provided by \citet{gloeckle2024mtp} in order to achieve best possible performance with minimal resource requirements.

We consider models of two sizes: 1.3B and 3B. We train our models on both subword and byte tokens. Byte-level models operate directly on raw bytes (e.g., UTF-8 encoded text), enabling them to handle diverse character sets and misspellings more robustly, while also reducing vocabulary size and memory usage \cite{byt5}.
The resulting vocabulary sizes, or model output dimensionality, for each type of models are 32,000 and 320 respectively. In order to maintain relative similarity in the amount of training steps that models undergo during the pre-training stage, we limit the context window size of subword-level models to 1024 and that of byte-level models to 4096.

\subsubsection{Prediction Heads}

As for the definition of additional language modeling heads, we consider the following two setups in our experiments:

\begin{itemize}[leftmargin=\parindent]
    \item \textbf{Linear Layers (LL)}: We add additional $k-1$ output linear layers with $ hidden\_{size} \times vocab\_{size} $ dimensions to a regular NTP architecture in case of a $k$-token prediction model. This essentially translates to linear probing of the model's hidden states. 
    
    While this setup allows for predictions that can be obtained in parallel, and thus not leading to any computational overheads with regards to output latency, it introduces an additional memory overhead. For subword-level models, this leads to additional 65M parameters per head in case of a 1.3B model, and additional 98M parameters per head in case of a 3B model. For byte-level models, this setup translates to only 0.6M and 1M additional trainable parameters per head respectively.

    \item \textbf{Transformer Layers (TL)}: We use $l-k$ transformer layers as the model backbone and dedicate $k$ out of $l$ layers to handling token prediction in the latent space before passing them through the shared output linear layer. 

    This setup not only does not introduce any additional trainable parameters, but also reduces the parameter count of the shared backbone by 51M (1.3B model) or 113M (3B model).
\end{itemize}

\subsubsection{Dataset} We conduct our experiments on the MiniPile dataset \cite{minipile}. It is a subset of The Pile corpus  \cite{the_pile}\footnote{Available under MIT License} that consists of 1M documents in the training split and 10K documents in the test split.

We train a dedicated BPE subword tokenizer \cite{bpe} with a vocabulary size of 32K on this dataset, which yields about 1.7B subword tokens for training. The corresponding training byte count is around 5.9B.

An important note regarding the discussion of our results is the fact that our models cannot be directly compared to other publicly released LLMs, given the size of the dataset. Our goal was not to train competitive LLMs, but rather to evaluate the viability of our proposed pre-training curricula. 

In order to confirm that our results hold true as the scale increases, we trained additional 1.3B MTP models with 4 TL heads together with a baseline 1.3B NTP model on 10B tokens of FineWeb-Edu \cite{fineweb-edu}. The performance of these models is shown in Table \ref{tab:10b_results} of the Appendix \ref{sec:appendix_res}.

\subsubsection{Training setup \& Hyperparameters}

We apply Best Fit Bin Packing \cite{best_fit_packing} to the train split of the MiniPile in order to prepare model inputs for the pre-training stage. We then train all of our models for 1 epoch. The training batch size is 1024. This setup allows us to iterate quickly over various model configurations.

In all experiments we schedule the learning rates with a linear warmup lasting 10\% of total training steps and cosine decay \cite{loshchilov2017sgdr} to 10\% of the peak learning rate, which is $2e-4$ for all models. All experiments use the AdamW optimizer \cite{loshchilov2018decoupled} with $\beta_1 = 0.9$, $\beta_2 = 0.95$ and decoupled $L_2$ weight decay coefficient of $0.1$. We also clip gradients to a maximal Euclidean norm of $1.0$ in all experiments. 

We plan to release the code used to train the models under our proposed curricula in our Github repository\footnote{\url{https://github.com/aynetdia/mtp_curriculum}}.

\begin{table*}[ht]
\resizebox{1\linewidth}{!}{
\begin{tabular}{ccc|ccccc|ccccc}
\toprule
\toprule
 &  &  & \multicolumn{5}{c|}{1.3B} & \multicolumn{5}{c}{3B} \\
\cline{4-13}
Tokens & Heads & Curriculum & MiniPile & LAMBADA & BLiMP & ARC-E & OBQA & MiniPile & LAMBADA & BLiMP & ARC-E & OBQA \\
 &  &  & BPB $\downarrow$ & BPB $\downarrow$ & Acc. $\uparrow$ & Acc. $\uparrow$ & Acc. $\uparrow$ & BPB $\downarrow$ & BPB $\downarrow$ & Acc. $\uparrow$ & Acc. $\uparrow$ & Acc. $\uparrow$ \\
\midrule
\midrule
\multirow[c]{13}{*}{Subword} & 1 LL & - & 1.08 & 1.34 & 71.80 & 34.60 & 25.80 & 1.03 & 1.17 & 74.93 & 34.60 & 24.80 \\
\cline{2-13}
 & \multirow[c]{3}{*}{2 LL} & - & 1.12 & 1.44 & 70.27 & 31.99 & 23.80 & 1.06 & 1.27 & 72.09 & 36.36 & 24.60 \\
 &  & Forward & \dab{+0.15\%} & \dab{+0.37\%} & \uag{+0.75\%} & \uag{+8.29\%} & \uag{+7.56\%} & \dab{+0.46\%} & \dab{+0.01\%} & \uag{+2.16\%} & \dab{-2.66\%} & \uag{+3.25\%} \\
 &  & Reverse & \uag{-2.19\%} & \uag{-5.79\%} & \uag{+1.69\%} & \uag{+1.84\%} & \uag{+5.04\%} & \uag{-1.78\%} & \uag{-4.91\%} & \uag{+3.60\%} & \dab{-0.23\%} & \dab{-1.62\%} \\
\cline{2-13}
 & \multirow[c]{3}{*}{2 TL} & - & 1.11 & 1.41 & 71.04 & 34.09 & 26.20 & 1.05 & 1.24 & 73.32 & 36.15 & 25.00 \\
 &  & Forward & \dab{+0.49\%} & \dab{+1.82\%} & \dab{-0.45\%} & \uag{+3.09\%} & \dab{-4.58\%} & \dab{+0.88\%} & \dab{+0.26\%} & \dab{-0.08\%} & \dab{-1.98\%} & \dab{-1.60\%} \\
 &  & Reverse & \uag{-1.98\%} & \uag{-4.19\%} & \uag{+0.88\%} & \uag{+2.10\%} & 0.00\% & \uag{-1.75\%} & \uag{-3.75\%} & \uag{+0.83\%} & \dab{-3.03\%} & \uag{+0.80\%} \\
\cline{2-13}
 & \multirow[c]{3}{*}{4 LL} & - & 1.19 & 1.61 & 67.48 & 34.22 & 24.40 & 1.12 & 1.44 & 72.20 & 36.45 & 25.00 \\
&  & Forward & \uag{-2.46\%} & \uag{-3.54\%} & \uag{+3.07\%} & \uag{+3.81\%} & \uag{+4.92\%} & \uag{-1.76\%} & \uag{-4.85\%} & \uag{+0.22\%} & \dab{-2.89\%} & \uag{+1.60\%} \\
 &  & Reverse & \uag{-5.99\%} & \uag{-13.32\%} & \uag{+3.35\%} & \uag{+1.97\%} & 0.00\% & \uag{-5.19\%} & \uag{-12.88\%} & \uag{+2.08\%} & \dab{-3.00\%} & \dab{-4.00\%} \\
\cline{2-13}
 & \multirow[c]{3}{*}{4 TL} & - & 1.18 & 1.57 & 67.07 & 33.96 & 23.40 & 1.11 & 1.38 & 71.34 & 36.41 & 23.20 \\
 &  & Forward & \uag{-2.39\%} & \uag{-4.37\%} & \uag{+2.16\%} & \uag{+0.87\%} & 0.00\% & \uag{-1.42\%} & \uag{-4.46\%} & \uag{+0.77\%} & \dab{-3.01\%} & \uag{+9.48\%} \\
 &  & Reverse & \uag{-5.52\%} & \uag{-9.85\%} & \uag{+4.58\%} & \uag{+0.74\%} & \uag{+1.71\%} & \uag{-4.84\%} & \uag{-9.61\%} & \uag{+2.63\%} & \uag{+1.62\%} & \dab{-4.31\%} \\
\midrule
\midrule
\multirow[c]{7}{*}{Byte} & 1 LL & - & 1.14 & 1.06 & 66.97 & 29.97 & 24.00 & 1.07 & 1.00 & 71.20 & 29.92 & 26.40 \\
\cline{2-13}
 & \multirow[c]{3}{*}{4 LL} & - & 1.16 & 1.09 & 69.13 & 29.42 & 30.40 & 1.08 & 1.00 & 71.42 & 29.97 & 30.60 \\
 &  & Forward & \dab{+3.16\%} & \dab{+2.82\%} & \dab{-2.76\%} & \dab{-1.43\%} & \dab{-6.58\%} & \dab{+2.62\%} & \dab{+3.12\%} & \dab{-1.40\%} & 0.00\% & \uag{+1.96\%} \\
 &  & Reverse & \uag{-3.44\%} & \uag{-3.95\%} & \uag{+1.77\%} & \dab{-0.72\%} & \uag{+6.58\%} & \uag{-2.43\%} & \uag{-4.45\%} & \uag{+1.46\%} & \uag{+7.30\%} & \uag{+1.31\%} \\
\cline{2-13}
 & \multirow[c]{3}{*}{8 LL} & - & 1.23 & 1.19 & 68.16 & 28.79 & 34.40 & 1.13 & 1.07 & 69.60 & 29.71 & 30.40 \\
 &  & Forward & \dab{+0.81\%} & 0.00\% & \dab{-4.41\%} & \dab{-1.17\%} & \dab{-19.77\%} & \dab{+0.79\%} & \dab{+1.28\%} & \uag{+0.83\%} & \uag{+2.27\%} & \dab{-7.24\%} \\
 &  & Reverse & \uag{-6.98\%} & \uag{-9.26\%} & \uag{+1.75\%} & \dab{-1.02\%} & \dab{-11.63\%} & \uag{-5.30\%} & \uag{-7.35\%} & \uag{+3.46\%} & \uag{+3.26\%} & \dab{-5.92\%} \\
\bottomrule
\bottomrule
\end{tabular}
}
\caption{Performance of MTP models on NTP tasks using only their main LM heads. Percentage values indicate relative \uag{improvements} or \dab{degradations} in benchmark scores compared to the respective static MTP baselines trained without a curriculum. 1 LL head models refer to standard NTP models. The same table with absolute scores is provided in Appendix \ref{sec:appendix_res}.}
\label{tab:tf_main_head}
\end{table*}

\subsection{Experiment 1: Main LM Head Performance}

\subsubsection{Setup} 

In this experiment we evaluate the model's NTP performance on the test set of the MiniPile dataset and on the OpenAI's version of the LAMBADA benchmark \cite{lambada}. We report the bits-per-byte metric instead of token perplexities, to allow for a direct comparison between subword- and byte-level models in our experiments. We use EleutherAI's evaluation harness (v0.4.3) \cite{eval-harness} to run our models on the benchmarks.

Since we are training our models on a significantly limited amount of data compared to what LLMs are usually trained on, we believe that widely used knowledge-based tasks like e.g.\ \noindent MMLU \cite{mmlu} would not be particularly meaningful to analyze their performance, as they training dataset does not cover an extended amount of factual knowledge. Thus we opt to focus on the BLiMP benchmark \cite{blimp}, since it allows for a fairer evaluation of the linguistic capabilities of our models. 

Nevertheless, we also include the evaluation on the easy set of the ARC challenge (ARC-E) \cite{arc_challenge} and on the OpenBookQA (OBQA) benchmark \cite{openbookqa} for the sake of a more comprehensive evaluation, as they focus on commonsense knowledge and reasoning, rather than area-specific knowledge. However we would like to note that these benchmarks do not provide an assessment of the underlying predictive capabilities associated with predicting multiple next tokens.

\subsubsection{Results} 

The performances\footnote{We also include a detailed breakdown of how our models perform on each linguistic phenomenon considered in the BLiMP benchmark individually in Tables \ref{tab:blimp_breakdown_1b} and \ref{tab:blimp_breakdown_3b} of Appendix \ref{sec:appendix_res}. This offers additional insight into the impact of various MTP objectives on how the models learn the underlying language dynamics.} of all models on the NTP tasks using only the main language modeling heads are listed in Table \ref{tab:tf_main_head}.
We make the following observations:

\noindent
\textbf{Byte-level dynamic MTP models outperform subword-level ones.} Despite starting from or converging to the NTP task during the pre-training, our proposed MTP curricula do not lead to performance improvements over the NTP baseline when applied to subword-level models of both sizes. 
However they lead to a noticeable improvement over the models trained with a static MTP objective, e.g. when they have 4 output heads, no matter if they are in the form of linear layers or transformer layers.
Byte-level Reverse Curriculum MTP models, on the other hand, are able to score on par with or even outperform both static MTP and NTP models in non-knowledge-based benchmarks. 

This difference between subword- and byte-level models can be explained by the fact that smaller LLMs are constrained by how much future token information they can retain and leverage at each prediction step due to their parameter count. And given that subword tokens carry more semantic and morphological information than a single byte, modeling multiple bytes at once is much easier than multiple subword tokens. We argue that the Reverse Curriculum MTP objective leads to richer hidden state representations by making the model predict multiple future bytes at once. This, in turn, improves performance on the NTP task, especially compared to subword-level SLMs, when the model is trained to look ahead up to 4 future tokens during pre-training.

\noindent \textbf{Performance is similar for LL and TL heads.} As for the performance difference between MTP models with multiple linear layers and transformer layers as language modeling heads, it is rather small, or sometimes even negligible, across both model sizes and both dynamic and static MTP objectives. And that is despite the fact that performing NTP with only the first LM heads results in Transformer Layer models "losing" parameters during inference. The effective parameter delta between a Transformer Layer MTP model with only the first LM head engaged and the corresponding Linear Layer model is around 210M per head in case of 3B model.

\begin{figure*}[ht]
  \includegraphics[width=\linewidth]{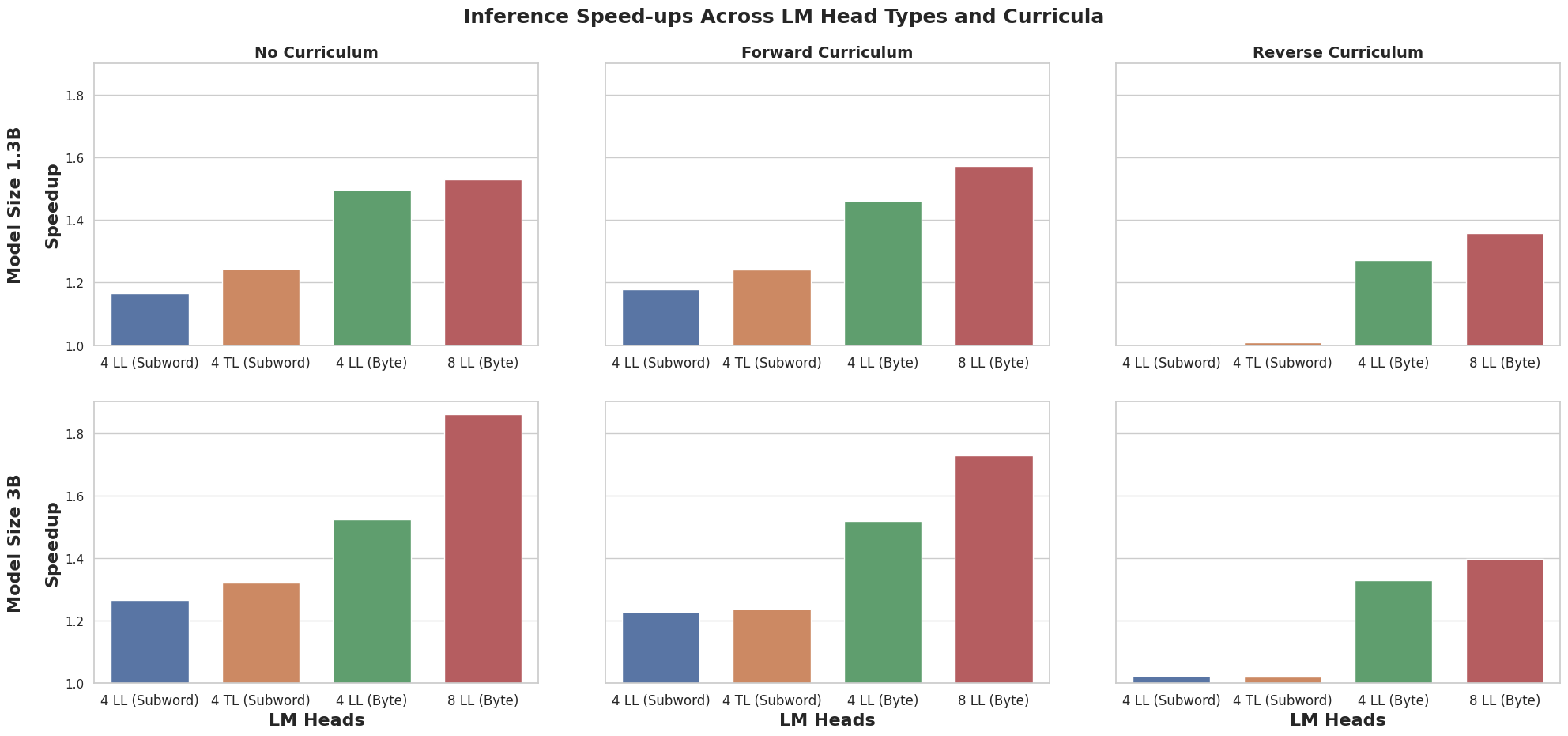} 
  \caption{Inference speed-ups achieved by MTP models in relation to the NTP baseline. We measure the speed-ups in terms of the amount of forward passes required to generate a sequence using self-speculative decoding formulation proposed by \citet{blockwise_parallel_decoding}}
  \label{fig:speedups}
\end{figure*}

\subsection{Experiment 2: Inference speed}

\subsubsection{Setup} 
We implement greedy blockwise speculative decoding \cite{blockwise_parallel_decoding} and measure decoding speeds on completing inputs from a the MiniPile test split. We sample 512 documents, take the first 256 subword or 512 byte tokens as inputs and generate completions of the same length. All completions are generated with batch size 8. We measure the speedups in terms of the amount of forward passes relative to the NTP baseline.

\begin{figure}[!h]
  \includegraphics[width=\columnwidth]{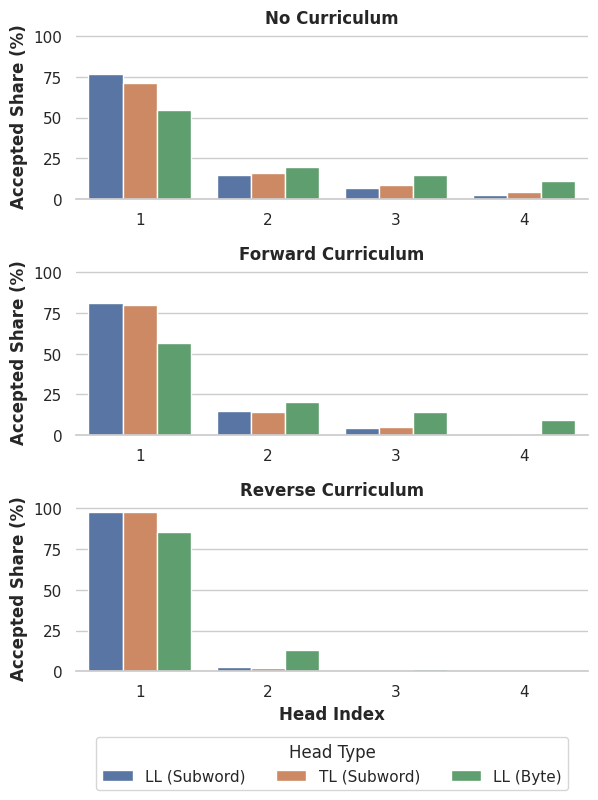} 
  \caption{Breakdown of the per-head acceptance rate of tokens drafted via self-speculative decoding for 3B subword models. The y-axis indicates a head's share of tokens accepted by the model in total.}
  \label{fig:speculative_accepted}
\end{figure}

\subsubsection{Results} 
Figure \ref{fig:speedups} showcases the maximum inference speed-ups that can be achieved by our MTP models, i.e. speed-ups when using all of the trained LM heads. Since self-speculative decoding requires at least two forward passes to predict and verify $k > 1$ tokens, 2-token subword models were excluded from this figure, as they do not allow for any inference speed-ups over a regular NTP setup.

While the static MTP objective results in the largest inference speed-ups across both model types, sizes, and LM head types, the Forward Curriculum comes as a very close second. Reverse Curriculum, on the other hand, basically converges towards NTP inference speeds regardless of the LM head type used with subword-level models, and offers only modest speed-ups in case of byte-level models.

\begin{table*}[ht]
\resizebox{1\linewidth}{!}{
\begin{tabular}{ccc|ccccc|ccccc}
\toprule
\toprule
 &  &  & \multicolumn{5}{c|}{1.3B} & \multicolumn{5}{c}{3B} \\
 \cline{4-13}
Tokens & Heads & Curriculum & BLEU $\uparrow$ & ROUGE-L $\uparrow$ & SemScore $\uparrow$ & TTR $\uparrow$ & G-Eval $\uparrow$ & BLEU $\uparrow$ & ROUGE-L $\uparrow$ & SemScore $\uparrow$ & TTR $\uparrow$ & G-Eval $\uparrow$ \\
\midrule
\midrule
\multirow[c]{13}{*}{Subword} & 1 LL & -  & 4.66 & 14.84 & 44.48 & 14.94 & 1.72 & 5.15 & 15.76 & 45.88 & 15.75 & 1.79 \\
\cline{2-13}
 & \multirow[c]{3}{*}{2 LL} & -  & 4.31 & 14.83 & 42.05 & 13.78 & 1.72 & 4.82 & 15.50 & 45.15 & 14.92 & 1.78 \\
 &  & Forward & \dab{-2.70\%} & \dab{-0.06\%} & \uag{+0.13\%} & \dab{-0.34\%} & \uag{+1.17\%} & \uag{+4.10\%} & \uag{+2.01\%} & \uag{+0.47\%} & \uag{+3.65\%} & \dab{-0.72\%} \\
 &  & Reverse & \uag{+7.21\%} & \uag{+0.46\%} & \uag{+3.01\%} & \uag{+7.46\%} & \uag{+1.42\%} & \uag{+1.25\%} & \uag{+1.35\%} & \uag{+1.24\%} & \uag{+8.78\%} & \uag{+0.58\%} \\
\cline{2-13}
 & \multirow[c]{3}{*}{2 TL} & -  & 4.31 & 14.64 & 42.01 & 13.70 & 1.71 & 5.08 & 15.65 & 45.35 & 15.60 & 1.79 \\
 &  & Forward & \uag{+0.94\%} & \uag{+1.29\%} & \dab{-0.99\%} & \uag{+3.18\%} & \uag{+0.50\%} & \dab{-7.44\%} & \dab{-3.97\%} & \dab{-2.00\%} & \dab{-10.34\%} & \dab{-3.41\%} \\
 &  & Reverse & \uag{+4.72\%} & \uag{+2.81\%} & \uag{+1.76\%} & \uag{+1.68\%} & \dab{-2.44\%} & \uag{+3.88\%} & \uag{+1.29\%} & \dab{-0.74\%} & \dab{-3.06\%} & \uag{+1.40\%} \\
\cline{2-13}
 & \multirow[c]{3}{*}{4 LL} & -  & 3.96 & 13.85 & 39.01 & 14.58 & 1.68 & 4.42 & 14.54 & 42.21 & 15.64 & 1.76 \\
 &  & Forward & \uag{+4.05\%} & \uag{+3.63\%} & \uag{+2.23\%} & \dab{-1.22\%} & \uag{+0.25\%} & \uag{+4.45\%} & \uag{+2.95\%} & \uag{+2.62\%} & \uag{+1.19\%} & \uag{+0.63\%} \\
 &  & Reverse & \uag{+15.80\%} & \uag{+7.50\%} & \uag{+8.77\%} & \uag{+9.26\%} & \uag{+2.00\%} & \uag{+13.04\%} & \uag{+6.46\%} & \uag{+5.34\%} & \uag{+0.03\%} & \uag{+1.98\%} \\
\cline{2-13}
 & \multirow[c]{3}{*}{4 TL} & -  & 4.24 & 14.11 & 39.02 & 14.39 & 1.69 & 4.78 & 14.87 & 43.17 & 15.78 & 1.78 \\
 &  & Forward & \dab{-8.87\%} & \dab{-0.14\%} & \uag{+3.07\%} & \uag{+1.19\%} & \dab{-0.21\%} & \dab{-3.83\%} & \uag{+1.16\%} & \uag{+1.44\%} & \uag{+2.91\%} & \uag{+0.14\%} \\
 &  & Reverse & \uag{+6.34\%} & \uag{+5.50\%} & \uag{+8.51\%} & \uag{+4.11\%} & \uag{+1.37\%} & \uag{+1.11\%} & \uag{+7.57\%} & \uag{+5.32\%} & \dab{-0.79\%} & \uag{+1.74\%} \\
\midrule
\midrule
\multirow[c]{7}{*}{Byte} & 1 LL & -  & 4.98 & 15.08 & 43.34 & 21.50 & 1.76 & 5.89 & 16.57 & 44.64 & 23.43 & 1.90 \\
\cline{2-13}
 & \multirow[c]{3}{*}{4 LL} & -  & 4.89 & 14.77 & 41.11 & 29.02 & 1.83 & 6.29 & 16.70 & 44.73 & 30.60 & 1.97 \\
 &  & Forward & \uag{+3.79\%} & \uag{+0.81\%} & \uag{+0.53\%} & \dab{-1.99\%} & \uag{+1.63\%} & \dab{-6.98\%} & \dab{-4.82\%} & \dab{-5.88\%} & \dab{-4.87\%} & \dab{-2.37\%} \\
 &  & Reverse & \uag{+0.03\%} & \uag{+5.88\%} & \uag{+0.92\%} & \uag{+9.45\%} & \uag{+1.68\%} & \dab{-5.51\%} & \dab{-3.21\%} & \dab{-0.12\%} & \uag{+4.51\%} & \dab{-0.41\%} \\
\cline{2-13}
 & \multirow[c]{3}{*}{8 LL} & -  & 4.77 & 14.94 & 39.91 & 35.49 & 1.86 & 5.41 & 15.78 & 41.79 & 41.70 & 1.93 \\
 &  & Forward & \uag{+2.80\%} & \dab{-0.48\%} & \dab{-5.20\%} & \dab{-0.51\%} & \dab{-2.91\%} & \uag{+4.65\%} & \uag{+1.21\%} & \uag{+0.59\%} & \dab{-12.72\%} & \uag{+0.14\%} \\
 &  & Reverse & \uag{+1.38\%} & \uag{+1.88\%} & \uag{+2.37\%} & \dab{-8.44\%} & \uag{+1.03\%} & \uag{+11.97\%} & \uag{+2.96\%} & \uag{+4.20\%} & \dab{-18.31\%} & \uag{+2.53\%} \\
\bottomrule
\bottomrule
\end{tabular}
}
\caption{Evaluation of model outputs on the MiniPile test set. Percentage values indicate relative \uag{improvements} or \dab{degradations} in evaluation metrics compared to the respective static MTP baselines trained without a curriculum. 1 LL models correspond to regular NTP models and their scores provided for reference. The same table with absolute scores is provided in Appendix \ref{sec:appendix_res}.}
\vspace{-3mm}
\label{tab:output_quality}
\end{table*} 

Figure \ref{fig:speculative_accepted} provides a per head breakdown of the self-speculative performance for every curriculum based on 3B models with 4 LM heads. Once again, there is a small difference in the acceptance rate of the tokens drafted by the additional heads between models trained with No and Forward curricula regardless of the head type, while almost no tokens drafted by additional heads trained under the Reverse Curriculum were accepted.

\subsection{Experiment 3: Output Quality}
\label{sec:output_quality}

\subsubsection{Setup}
We take the outputs generated by the models when performing Experiment 2 and compare them with "gold" completions of the same length from the MiniPile test set. We rely on BLEU \cite{bleu}, ROUGE-L \cite{rouge} and SemScore \cite{semscore} metrics to determine how closely the model completions match the ground truth from the MiniPile test set. We also calculate the Type-Token Ratio (TTR) metric, which is a measure of lexical diversity. It is defined as the ratio of unique words to the total number of words in a text. 

We use the NLTK's (v.3.8.2) \cite{nltk} implementation of the BLEU score, as well as the RegExp tokenizer to tokenize the generated and gold sequences for BLEU, ROUGE-L and TTR. We provide the regular expression used to tokenize the text sequences in Appendix \ref{sec:appendix_eval_details}. For SemScore we use the \texttt{all-mpnet-base-v2} model \cite{sent_transformers}. As for the ROUGE-L, we use Google's implementation of it.

In addition to the aforementioned "traditional" metrics, we also perform an LLM-based (LLM-as-a-judge) evaluation of generated outputs. We will refer to the resulting metric as G-Eval, as we rely on \citet{g-eval} for the prompt design and probability-weighted final score calculation. Our evaluation prompt can be found in Appendix \ref{sec:appendix_eval_details}. We define a 5-point rating scale with 1 being the worst score, and 5 being the best loosely based on \citet{self-instruct}. The model of our choice for the role of an evaluator is OpenAI's \texttt{gpt-3.5-turbo-0125} due to its cost efficiency.

\subsubsection{Results} 

The results of evaluation are listed in Table \ref{tab:output_quality}. We also include the results for NTP models for reference.
We make the following observations:

\noindent
\textbf{Proposed curricula outperform static MTP baselines.} In almost all of the model configurations both of our proposed curricula lead to better results than the static MTP approach on at least 2 metrics. The only model configurations in which No Curriculum models achieve better results are a byte-level 3B model with 4 linear layer heads and a subword-level 3B model with 2 transformer layer heads.

Even though the Reverse Curriculum often showcases greater quality of improvements than Forward Curriculum over No curriculum, we have already shown previously that the Reverse Curriculum yields models that are closer in nature to NTP models, both in terms of the performance of the main LM head, as well as the lack of self-speculative decoding speed-ups. Therefore it would be more appropriate to compare Reverse Curriculum MTP models with NTP models, rather than No Curriculum MTP models. In this regard, the Reverse Curriculum models are not able to generate outputs that are qualitatively better than NTP models.

When comparing No Curriculum models only against the Forward Curriculum MTP models, the latter still come out on top on at least two metrics, but with two additional exceptions: byte-level model with 1.3B parameters and 8 Linear Layer heads, as well as a subword-level model with 1.3B parameters and 2 Linear Layer heads. Nevertheless, in 8 out of 12 different configurations in total Forward Curriculum MTP generate qualitatively better sequences with almost the same speed-ups via self-speculative decoding as MTP models with No Curriculum.

\noindent \textbf{TL heads outperform LL.} Additional model parameters that are introduced by using auxiliary linear layers as LM heads do not reliably translate to better generated sequences of better quality. Even the subword-level 3B models with 4 LM heads showcase better output quality with Transformer Layer LM Heads, rather than Linear Layers.

This further reinforces the idea that MTP is not simply a matter of the parameter count. Linear Layers on their own might be too simplistic as LM heads to enable better MTP. 

\noindent \textbf{Byte-level MTP models outperform subword models across all metrics.} In Experiment 1 we already argued that by modeling multiple bytes at once, byte-level MTP models arrive at hidden representations that can handle more complex patterns, compared to their NTP counterparts. The output quality evaluation of 3B byte-level models further reinforces this point, and interestingly enough the MTP models generate significantly more lexically diverse outputs, as indicated by the TTR metric. 

In addition, byte-level models beat subword-level models across all metrics and model configurations. This also holds true not only when comparing byte-level MTP models against subword-level MTP counterparts, but also when comparing both static and dynamic MTP models against the subword NTP baseline.

\section{Related Work}

\noindent \textbf{Curriculum learning.} After \citet{bengio2009curriculum} first proposed to apply a curriculum learning strategy in the context of machine learning, it has been successfully applied on a number of tasks in various machine learning domains including natural language processing, computer vision and speech recognition \cite{cl_survey}. In the context of language modeling, they have been shown to provide benefits both when pre-training encoder-only models \cite{cl_nlu, cl_bert, bert_lrc}, as well as instruction-tuning large decoder-only models \cite{orca, curr_instr}. 

The use of curriculum learning approaches was not reported in pre-training any publicly available decoder-only foundation models trained on vast amounts of text data, although recently \citet{feng2024} showed that using a two-stage curriculum based on text quality can lead better training outcomes. Meanwhile curriculum learning approaches have been very popular in data-constrained pre-training setups \cite{babylm_2023, babylm_2024}. While the curricula that focus on ordering the data based on various difficulty metrics were not found to be consistently better than non-curriculum baselines, an approach by \citet{less_is_more} that involves a curriculum for pre-training objectives was able to reliably outperform non-curriculum baselines in a data-constrained setup.

\noindent \textbf{Multi-token prediction.} ProphetNet \cite{prophetnet} was the first large-scale transformer-based language model that was able to predict multiple n-grams in one prediction step. 
However, their model relies on n-stream self-attention mechanism that involves more computational overhead compared to regular transformers.

\citet{future_lens} showed that hidden states of next-token prediction models are able to encode more than a single token ahead by probing pre-trained transformers, and that it's possible to predict those to a certain extent.

\citet{gloeckle2024mtp} improved upon the previous work by proposing slight architectural tweaks, such as using full transformer layers as language modeling heads, to account for the multi-token prediction task that resulted in a more computationally efficient, compute-matched with NTP models, and effective method for multi-token prediction.

\noindent \textbf{Self-speculative decoding.} \citet{blockwise_parallel_decoding} were the first to suggest a speculative decoding scheme for faster inference. Since then, a number self-speculative decoding methods were introduced. Some of these methods rely on the early-exit mechanism \cite{layerskip, kangaroo}, others on skipping intermediate layers \cite{draft_verify, swift}, and some on architectural transformations \cite{koala}. Medusa \cite{medusa} has gained the most prominence due to its simplicity and ability to relatively easily and cost-efficiently enable self-speculative decoding for LLMs that were pre-trained using the regular NTP objective.

\section{Conclusion}

This paper introduces a novel curriculum-based training strategy to the multi-token prediction task. We conducted extensive experiments to determine their validity against both the regular NTP, as well as the static MTP training objectives. Our experiments demonstrated that the Forward MTP Curriculum offers the best trade-offs between the main LM head performance, inference speed, as well as the final output quality, when compared against both the static MTP and Reverse MTP Curriculum approaches. As a result, it allows SLMs to better leverage the MTP objective during pre-training. 

We have also shown that Reverse MTP Curriculum models outperform static MTP models on NTP benchmarks. In addition, Reverse MTP Curriculum models generate sequences of better quality than NTP models in case of both subword- and byte-level models. Unfortunately, they are not able to offer any meaningful inference speed-ups via self-speculative decoding, while coming short on NTP tasks in comparison to models trained using solely the NTP objective.

In the future we are interested in investigating the potential benefits of the MTP objective for non-transformer architectures.

\section*{Limitations}

One limitation of our proposed curricula is the fact that they are pre-defined in advance. The decision to progressively add or remove a token to or from an $m$-token task every $\frac{m}{n}$ steps when training for $n$ steps is somewhat arbitrary, since it does not rely on any metrics about the the models themselves or the training loss. This may lead to a situation that perhaps some LM heads require more steps to be fully "saturated", while others require less, and not accounting for this results in under- or overfitted LM heads. Since the goal of this paper was to establish that curriculum-based MTP training fundamentally benefits small language models over a static MTP objective in principle, we leave exploring the potentially very large space of possible schedules for future work.

In addition, we acknowledge the fact that the language models considered in our experiments were trained on a dataset that is significantly smaller in size than datasets on which other contemporary LLMs are trained. We do not exclude the possibility that, when trained for a prolonged time and on significantly more data, the evaluation results of our proposed curricula may differ from the results reported in our paper in one way or another. However given the consistency in the results across various model configurations in all of our experiments, we argue that the training dataset size used to determine the validity of our proposed curricula does not undermine the reliability of our conclusions. 


\section*{Acknowledgments}

We thank all reviewers for their valuable feedback. The authors are supported by the Deutsche Forschungsgemeinschaft (DFG, German Research Foundation) under Emmy Noether grant “Eidetic Representations of Natural Language” (project number 448414230). Further, Alan Akbik is supported by the Deutsche Forschungsgemeinschaft (DFG, German Research Foundation) under Germany’s Excellence Strategy "Science of Intelligence" (EXC 2002/1, project number 390523135).

The authors gratefully acknowledge the scientific support and HPC resources provided by the Erlangen National High Performance Computing Center (NHR@FAU) of the Friedrich-Alexander-Universität Erlangen-Nürnberg (FAU) under the NHR project c106fa. NHR funding is provided by federal and Bavarian state authorities. NHR@FAU hardware is partially funded by the German Research Foundation (DFG) – 440719683.

\bibliography{main}

\newpage

\appendix

\section{Additional Evaluation Details}

\label{sec:appendix_eval_details}

\subsection{NLTK RegExp Tokenizer Rule}

Below we provide the regular expression rule assigned to a Python variable that we used for the NLTK's RegExp tokenizer. This tokenizer was used to tokenize the sequences generated by the models we have trained  in order to calculate the output quality metrics in Section \ref{sec:output_quality}, namely BLEU, ROUGE-L and TTR.

\begin{Verbatim}[fontsize=\small]
pattern = r'''
# strings
"(?:[^"\\]|\\.)*"                   |
'(?:[^'\\]|\\.)*'                   |
# numeric literals
\d+(?:\.\d+)?                       |
# words/unicode identifiers
[\w_]+                              |
# punctuation, brackets, etc.
[!@#\$%\^&\*\(\)\-=\+\\|\[\]\{\};:'",.<>/?`~]+
'''
\end{Verbatim}

\subsection{G-Eval Prompt}

\textit{You will be given a sequence triplet consisting of:}
\begin{enumerate}
    \item \textit{An input sequence (text or code) that serves as the starting point.}
    \item \textit{An output sequence written as a continuation of the input.}
    \item \textit{A target sequence that represents the expected continuation of the input sequence.}
\end{enumerate}

\noindent\textit{Your task is to rate the written output sequence on one metric.} 

\noindent\textit{Please make sure you read and understand these instructions carefully. Keep this document open while reviewing, and refer to it as needed.}

\vspace{2mm}
\noindent\textit{Evaluation Criteria:}
\vspace{2mm}

\noindent\textit{Overall Quality (1-5) - how well does the output sequence continue the input sequence and align with the target sequence?}
\begin{itemize}
    \item[-] \textit{A score of 5 means that the output sequence is excellent. It provides a seamless continuation of the input sequence, closely aligns with the target sequence, and avoids any repetitions, irrelevant passages, or major errors.}
    \item[-] \textit{A score of 4 means that the output sequence is good. It continues the input sequence well and mostly aligns with the target sequence, but may include minor errors or imperfections, such as slight incoherence or small structural issues.}
    \item[-] \textit{A score of 3 means that the output sequence is acceptable. It maintains some relevance to the input sequence and partial alignment with the target sequence, but contains noticeable flaws, such as incoherence, repetitions, or deviations that reduce its quality.}
    \item[-] \textit{A score of 2 means that the output sequence is poor. It struggles to continue the input sequence coherently or deviates significantly from the target sequence, with major errors, irrelevant sections, or repeated patterns.}
    \item[-] \textit{A score of 1 means that the output sequence is invalid. It fails to continue the input sequence meaningfully, shows no alignment with the target sequence, or is completely incoherent.}
\end{itemize}

\textit{Evaluation Steps:}
\begin{enumerate}
    \item \textit{Carefully read the input, output, and target sequences.}
    \item \textit{Compare the output sequence to both the input sequence (continuity) and the target sequence (alignment).}
    \item \textit{Rate the output on a scale of 1-5 for Quality, according to the criteria above.}
\end{enumerate}

\textit{\#\#\# Input Sequence:}

\textit{\{Input Sequence\}}

\textit{\#\#\# Output Sequence:}

\textit{\{Output Sequence\}}

\textit{\#\#\# Target Sequence:} 

\textit{\{Target Sequence\}}

\vspace{2mm}
\textit{Evaluation Form (scores ONLY):} 

\textit{- Quality:}


\begin{table*}[]
\resizebox{1\linewidth}{!}{
\begin{tabular}{ccc|ccccc|ccccc}
\toprule
\toprule
 &  &  & \multicolumn{5}{c|}{1.3B} & \multicolumn{5}{c}{3B} \\
\cline{4-13}
Tokens & Heads & Curriculum & MiniPile & LAMBADA & BLiMP & ARC-E & OBQA & MiniPile & LAMBADA & BLiMP & ARC-E & OBQA \\
 &  &  & BPB $\downarrow$ & BPB $\downarrow$ & Acc. $\uparrow$ & Acc. $\uparrow$ & Acc. $\uparrow$ & BPB $\downarrow$ & BPB $\downarrow$ & Acc. $\uparrow$ & Acc. $\uparrow$ & Acc. $\uparrow$ \\
\midrule
\midrule
\multirow[c]{13}{*}{Subword} & 1 LL & - & \textbf{1.08} & \textbf{1.34} & \textbf{71.80} & 34.60 & 25.80 & \textbf{1.03} & \textbf{1.17} & \textbf{74.93} & 34.60 & 24.80 \\
\cline{2-13}
 & \multirow[c]{3}{*}{2 LL} & - & 1.12 & 1.44 & 70.27 & 31.99 & 23.80 & 1.06 & 1.27 & 72.09 & 36.36 & 24.60 \\
 &  & Forward & 1.12 & 1.45 & 70.79 & 34.64 & 25.60 & 1.06 & 1.27 & 73.65 & 35.40 & \textbf{25.40} \\
 &  & Reverse & 1.09 & 1.36 & 71.45 & 32.58 & 25.00 & 1.04 & 1.20 & 74.69 & 36.28 & 24.20 \\
\cline{2-13}
 & \multirow[c]{3}{*}{2 TL} & - & 1.11 & 1.41 & 71.04 & 34.09 & \textbf{26.20} & 1.05 & 1.24 & 73.32 & 36.15 & 25.00 \\
 &  & Forward & 1.12 & 1.43 & 70.72 & 35.14 & 25.00 & 1.06 & 1.24 & 73.26 & 35.44 & 24.60 \\
 &  & Reverse & 1.09 & 1.35 & 71.66 & 34.81 & \textbf{26.20} & 1.03 & 1.19 & 73.93 & 35.06 & 25.20 \\
\cline{2-13}
 & \multirow[c]{3}{*}{4 LL} & - & 1.19 & 1.61 & 67.48 & 34.22 & 24.40 & 1.12 & 1.44 & 72.20 & 36.45 & 25.00 \\
 &  & Forward & 1.16 & 1.55 & 69.55 & \textbf{35.52} & 25.60 & 1.10 & 1.37 & 72.35 & 35.40 & \textbf{25.40} \\
 &  & Reverse & 1.12 & 1.43 & 69.74 & 34.89 & 24.40 & 1.06 & 1.26 & 73.70 & 35.35 & 24.00 \\
\cline{2-13}
 & \multirow[c]{3}{*}{4 TL} & - & 1.18 & 1.57 & 67.07 & 33.96 & 23.40 & 1.11 & 1.38 & 71.34 & 36.41 & 23.20 \\
 &  & Forward & 1.16 & 1.50 & 68.52 & 34.26 & 23.40 & 1.09 & 1.32 & 71.89 & 35.31 & \textbf{25.40} \\
 &  & Reverse & 1.12 & 1.42 & 70.14 & 34.22 & 23.80 & 1.06 & 1.25 & 73.21 & \textbf{36.99} & 22.20 \\
\midrule
\midrule
\multirow[c]{7}{*}{Byte} & 1 LL & - & 1.14 & 1.06 & 66.97 & \textbf{29.97} & 24.00 & 1.07 & 1.00 & 71.20 & 29.92 & 26.40 \\
\cline{2-13}
 & \multirow[c]{3}{*}{4 LL} & - & 1.16 & 1.09 & 69.13 & 29.42 & 30.40 & 1.08 & 1.00 & 71.42 & 29.97 & 30.60 \\
 &  & Forward & 1.20 & 1.13 & 67.22 & 29.00 & 28.40 & 1.11 & 1.03 & 70.43 & 29.97 & \textbf{31.20} \\
 &  & Reverse & \textbf{1.12} & \textbf{1.05} & \textbf{70.36} & 29.21 & 32.20 & \textbf{1.06} &\textbf{0.96} & \textbf{72.47} & \textbf{32.15} & 31.00 \\
\cline{2-13}
 & \multirow[c]{3}{*}{8 LL} & - & 1.23 & 1.19 & 68.16 & 28.79 & \textbf{34.40} & 1.13 & 1.07 & 69.60 & 29.71 & 30.40 \\
 &  & Forward & 1.24 & 1.19 & 65.15 & 28.45 & 27.60 & 1.14 & 1.08 & 70.18 & 30.39 & 28.20 \\
 &  & Reverse & 1.15 & 1.08 & 69.35 & 28.49 & 30.40 & 1.07 & 0.99 & 72.01 & 30.68 & 28.60 \\
\bottomrule
\bottomrule
\end{tabular}
}
\caption{Performance of MTP models using only their main LM heads on NTP tasks. Best NTP performance across all model configurations are \textbf{highlighted}.}
\label{tab:tf_main_head_abs}
\end{table*}

\begin{table*}[]
\resizebox{1\linewidth}{!}{
\begin{tabular}{ccc|ccccc|ccccc}
\toprule
\toprule
 &  &  & \multicolumn{5}{c|}{1.3B} & \multicolumn{5}{c}{3B} \\
 \cline{4-13}
Tokens & Heads & Curriculum & BLEU $\uparrow$ & ROUGE-L $\uparrow$ & SemScore $\uparrow$ & TTR $\uparrow$ & G-Eval $\uparrow$ & BLEU $\uparrow$ & ROUGE-L $\uparrow$ & SemScore $\uparrow$ & TTR $\uparrow$ & G-Eval $\uparrow$ \\
\midrule
\midrule
\multirow[c]{13}{*}{Subword} & 1 LL & - & 4.66 & 14.84 & 44.48 & 14.94 & 1.71 & 5.15 & 15.76 & 45.88 & 15.75 & 1.79 \\
\cline{2-13}
 & \multirow[c]{3}{*}{2 LL} & - & 4.31 & 14.83 & 42.05 & 13.78 & 1.72 & 4.82 & 15.50 & 45.15 & 14.92 & 1.78 \\
 &  & Forward & 4.20 & 14.82 & 42.11 & 13.74 & 1.74 & \textbf{5.02} & \textbf{15.81} & 45.36 & 15.47 & 1.77 \\
 &  & Reverse & \textbf{4.62} & \textbf{14.90} & \textbf{43.32} & \textbf{14.81} & \textbf{1.75} & 4.88 & 15.71 & \textbf{45.71} & \textbf{16.23} & \textbf{1.79} \\
\cline{2-13}
 & \multirow[c]{3}{*}{2 TL} & - & 4.31 & 14.64 & 42.01 & 13.70 & 1.71 & 5.08 & 15.65 & \textbf{45.35} & \textbf{15.60} & 1.79 \\
 &  & Forward & 4.35 & 14.83 & 41.59 & \textbf{14.14} & \textbf{1.72} & 4.70 & 15.03 & 44.44 & 13.98 & 1.72 \\
 &  & Reverse & \textbf{4.52} & \textbf{15.05} & \textbf{42.75} & 13.93 & 1.67 & \textbf{5.28} & \textbf{15.85} & 45.01 & 15.12 & \textbf{1.81} \\
\cline{2-13}
 & \multirow[c]{3}{*}{4 LL} & - & 3.96 & 13.85 & 39.01 & 14.58 & 1.68 & 4.42 & 14.54 & 42.21 & 15.64 & 1.76 \\
 &  & Forward & 4.12 & 14.36 & 39.88 & 14.40 & 1.69 & 4.61 & 14.97 & 43.31 & \textbf{15.83} & 1.77 \\
 &  & Reverse & \textbf{4.59} & \textbf{14.89} & \textbf{42.42} & \textbf{15.92} & \textbf{1.72} & \textbf{4.99} & \textbf{15.48} & \textbf{44.46} & 15.65 & \textbf{1.80} \\
\cline{2-13}
 & \multirow[c]{3}{*}{4 TL} & - & 4.24 & 14.11 & 39.02 & 14.39 & 1.69 & 4.78 & 14.87 & 43.17 & 15.78 & 1.78 \\
 &  & Forward & 3.86 & 14.09 & 40.21 & 14.56 & 1.68 & 4.59 & 15.04 & 43.79 & \textbf{16.23} & 1.78 \\
 &  & Reverse & \textbf{4.51} & \textbf{14.88} & \textbf{42.34} & \textbf{14.98} & \textbf{1.71} & \textbf{4.83} & \textbf{15.99} & \textbf{45.47} & 15.65 & \textbf{1.81} \\
\midrule
\midrule
\multirow[c]{7}{*}{Byte} & 1 LL & - & 4.98 & 15.08 & 43.34 & 21.50 & 1.76 & 5.89 & 16.57 & 44.64 & 23.43 & 1.90 \\
\cline{2-13}
 & \multirow[c]{3}{*}{4 LL} & - & 4.89 & 14.77 & 41.11 & 29.02 & 1.83 & 6.29 & \textbf{16.70} & \textbf{44.73} & 30.60 & \textbf{1.97} \\
 &  & Forward & \textbf{5.08} & 14.89 & 41.32 & 28.44 & \textbf{1.86} & 5.85 & 15.90 & 42.10 & 29.11 & 1.92 \\
 &  & Reverse & 4.89 & \textbf{15.64} & \textbf{41.48} & \textbf{31.76} & \textbf{1.86} & \textbf{5.94} & 16.17 & 44.68 & \textbf{31.98} & 1.96 \\
\cline{2-13}
 & \multirow[c]{3}{*}{8 LL} & - & 4.77 & 14.94 & 39.91 & \textbf{35.49} & 1.86 & 5.41 & 15.78 & 41.79 & \textbf{41.70} & 1.93 \\
 &  & Forward & \textbf{4.90} & 14.87 & 37.83 & 35.31 & 1.81 & 5.66 & 15.98 & 42.04 & 36.40 & 1.94 \\
 &  & Reverse & 4.83 & \textbf{15.22} & \textbf{40.85} & 32.50 & \textbf{1.88} & \textbf{6.06} & \textbf{16.25} & \textbf{43.54} & 34.07 & \textbf{1.98} \\
\bottomrule
\bottomrule
\end{tabular}
}
\caption{Evaluation of the outputs generated by the models based on the MiniPile test set inputs. \textbf{Highlighted} are scores that are improvements over the corresponding static MTP baselines trained without a curriculum for each respective model configuration.}
\label{tab:output_quality_abs}
\end{table*}

\section{Additional Results}
\label{sec:appendix_res}

\subsection{Result tables with absolute values}

Tables \ref{tab:tf_main_head_abs} and \ref{tab:output_quality_abs} report the absolute scores that correspond to those reported in Tables \ref{tab:tf_main_head} and \ref{tab:output_quality}.

\subsection{MTP models trained on 10B tokens}

\begin{table*}[ht]
\resizebox{1\linewidth}{!}{
\begin{tabular}{ccc|ccccc|ccccc}
\toprule
\toprule
 &  &  & \multicolumn{5}{c|}{Main LM Head} & \multicolumn{5}{c}{Output Quality} \\
\cline{4-13}
Tokens & Heads & Curriculum & MiniPile & LAMBADA & BLiMP & ARC-E & OBQA & \multirow[c]{2}{*}{BLEU $\uparrow$} & \multirow[c]{2}{*}{ROUGE-L $\uparrow$} & \multirow[c]{2}{*}{SemScore $\uparrow$} & \multirow[c]{2}{*}{TTR $\uparrow$} & \multirow[c]{2}{*}{G-Eval $\uparrow$} \\
 &  &  & PPL $\downarrow$ & PPL $\downarrow$ & Acc. $\uparrow$ & Acc. $\uparrow$ & Acc. $\uparrow$ &  &  &  &  & \\
\midrule
\midrule
\multirow[c]{3}{*}{Subword} & 1 LL & No & 47 & 74 & 78.74 & 54.63 & 30.40 & 2.02 & 15.39 & 52.40 & 13.18 & 1.52 \\
\cline{2-13}
 & \multirow[c]{3}{*}{4 TL} & No & 57 & 100 & 75.96 & 55.22 & 27.40 & 1.97 & 14.80 & 51.69 & 16.11 & 1.54 \\
 &  & Forward & 57 & 101 & 76.56 & 54.30 & 28.40 & 2.00 & 15.08 & 52.71 & 15.48 & 1.56 \\
 &  & Reverse & 50 & 75 & 77.55 & 55.60 & 30.40 & 2.19 & 14.91 & 52.77 & 16.25 & 1.56 \\
\bottomrule
\bottomrule
\end{tabular}
}
\caption{Evaluation results of 1.3B models trained on 10B tokens of FineWeb-Edu.}
\label{tab:10b_results}
\end{table*}

Table \ref{tab:10b_results} lists the evaluation results of 1.3B models trained on 10B subword tokens of FineWeb-Edu. The MTP models with 4 TL layers showcase a performance that is consistent with 4 TL MTP models trained on MiniPile. As the scale increases, the dynamic MTP objective continues to result in performance improvements over the static one. 

Granted, the NTP objective still leads to a best performing model, however we believe that as the model size increases, the static MTP objective should start outperforming the NTP objective, as shown by \citet{gloeckle2024mtp}. And given our observations, the dynamic MTP objective should result in even better performing scaled models.

We recognize that 10B tokens is still a very modest token budget for moder LLM pre-training. Nevertheless, we hope that our results will serve as a useful starting point for future work in a scaled setup.

\subsection{BLiMP breakdown}
\label{sec:blimp_breakdown}

Tables \ref{tab:blimp_breakdown_1b} and \ref{tab:blimp_breakdown_3b} break down the performance of the models on the BLiMP benchmark from Table \ref{tab:tf_main_head}. We show the performance on each linguistic phenomenon considered in the benchmark. Please refer to \citet{blimp} for the precise definition of each phenomenon and additional details on how to interpret them. Below we discuss the results of one of the particularly interesting and illustrative English phenomenon pair.

The category of tasks in BLiMP that fall under the Binding category deals with structural relationships between a pronoun and its antecedent. Often they are further apart from each other within a sentence, making them a good match for MTP models. Meanwhile, the Subject-Verb Agreement (SVA) phenomenon can be characterized as the agreement between subjects and verbs in number. Usually they are close to each other in a sentence, i.e. SVA often deals with relationships between adjacent tokens, which NTP models should excel at. 

The results show that even though the MTP models should be better suited to deal with the Binding phenomenon, the complexity of the MTP objective hinders the performance of 1.3B models with respect to this phenomenon. Only 3B MTP models are able to outperform their NTP counterparts, if trained with a dynamic MTP objective. Given the performance gains observed by 3B models with 4 LL or 2 TL heads, the performance degradation of 1.3B models in the Binding category can perhaps be explained by the limited capacity of the model due to its parameter count.

The performance gap on the SVA phenomenon, on the other hand, is more expected, as NTP models do not have to account for multiple future tokens at once in its internal representations. Notably, the performance drop is especially noticeable in case of 4-token prediction models. Notably, the dynamic MTP objective significantly alleviates the performance drop in the SVA category, especially when it comes to 3B models.

To conclude, these results make it evident that the proposed dynamic MTP objective offers some noticeable performance improvements in some categories of the English grammatical phenomena, despite the fact that MTP models on average perform worse than NTP models on the BLiMP benchmark, as shown in Tables \ref{tab:tf_main_head} or \ref{tab:tf_main_head_abs} of the paper.

\begin{table*}[ht]
\resizebox{1\linewidth}{!}{
\begin{tabular}{ccc|ccccccccccc}
\toprule
\toprule
 &  &  & Anaphor & Argument & Binding & Control/ & Ellipsis & Filler & Irregular & Island & NPI & Quantifiers & Subject-Verb \\
Tokens & Heads & Curriculum & Agreement & Structure &  & Raising &  & Gap & Forms & Effects & Licensing &  & Agreement  \\
\midrule
\midrule
\multirow[c]{13}{*}{Subword} & 1 LL & -  & 90.8 & 73.44 & 78.37 & 70.94 & 73.7 & 71.33 & 92.3 & 50.28 & 60.87 & 67.35 & 78.13 \\
\cline{2-14}
 & \multirow[c]{3}{*}{2 LL} & -  & 88.2 & 73.14 & 76.57 & 70.42 & 75.45 & 69.73 & 86.9 & 48.72 & 61.11 & 62.35 & 76.98 \\
 &  & Forward  & 88.35 & 72.91 & 76.87 & 69.54 & 74.8 & 69.27 & 93.15 & 53.59 & 61.21 & 54.52 & 78.37 \\
 &  & Reverse  & 87.8 & 74.51 & 77.35 & 71.06 & 75.6 & 70.06 & 87.15 & 49.57 & 63.21 & 64.65 & 78.6 \\
\cline{2-14}
 & \multirow[c]{3}{*}{2 TL} & -  & 89.75 & 72.59 & 76.95 & 68.34 & 71.95 & 69.94 & 91.65 & 50.55 & 62.69 & 67.35 & 77.42 \\
 &  & Forward  & 90.95 & 72.82 & 77.67 & 68.78 & 72.8 & 68.83 & 90.9 & 49.84 & 59.66 & 63.95 & 77.7 \\
 &  & Reverse  & 86.0 & 73.48 & 78.32 & 69.36 & 72.45 & 70.07 & 92.25 & 50.4 & 61.01 & 73.05 & 78.12 \\
\cline{2-14}
 & \multirow[c]{3}{*}{4 LL} & -  & 84.15 & 70.22 & 76.2 & 66.96 & 72.95 & 68.03 & 88.1 & 53.01 & 48.43 & 56.88 & 71.18 \\
 &  & Forward  & 88.1 & 71.43 & 77.37 & 68.26 & 70.65 & 69.49 & 91.45 & 51.64 & 57.06 & 56.12 & 76.82 \\
 &  & Reverse  & 85.95 & 73.19 & 77.78 & 68.38 & 73.7 & 68.4 & 89.25 & 52.21 & 55.17 & 62.05 & 74.15 \\
\cline{2-14}
 & \multirow[c]{3}{*}{4 TL} & -  & 90.5 & 69.19 & 75.82 & 67.02 & 71.4 & 68.67 & 83.4 & 48.59 & 52.73 & 57.73 & 70.33 \\
 &  & Forward  & 86.7 & 71.39 & 77.7 & 68.66 & 69.3 & 70.19 & 86.45 & 46.4 & 56.81 & 52.62 & 76.1 \\
 &  & Reverse  & 86.8 & 72.47 & 76.68 & 70.4 & 73.0 & 69.19 & 86.75 & 50.61 & 60.29 & 64.53 & 75.67 \\
\midrule
\midrule
\multirow[c]{7}{*}{Byte} & 1 LL & -  & 77.05 & 71.8 & 75.27 & 67.64 & 72.9 & 67.81 & 87.7 & 45.53 & 40.89 & 64.2 & 77.92 \\
\cline{2-14}
 & \multirow[c]{3}{*}{4 LL} & -  & 72.5 & 73.78 & 76.77 & 67.82 & 75.7 & 67.0 & 85.45 & 47.09 & 50.26 & 66.8 & 82.45 \\
 &  & Forward  & 71.65 & 71.12 & 75.0 & 64.54 & 71.5 & 67.0 & 82.35 & 44.61 & 57.39 & 57.38 & 75.12 \\
 &  & Reverse  & 72.75 & 73.78 & 75.03 & 67.98 & 74.55 & 69.94 & 87.3 & 47.7 & 57.03 & 65.83 & 84.18 \\
\cline{2-14}
 & \multirow[c]{3}{*}{8 LL} & -  & 69.15 & 72.04 & 75.7 & 65.78 & 74.9 & 67.9 & 88.25 & 46.32 & 54.96 & 68.2 & 76.4 \\
 &  & Forward  & 64.75 & 68.79 & 76.53 & 65.72 & 72.7 & 66.01 & 82.55 & 42.61 & 45.39 & 67.1 & 71.25 \\
 &  & Reverse  & 75.9 & 74.6 & 76.2 & 66.86 & 71.8 & 69.9 & 88.4 & 48.24 & 49.06 & 69.8 & 77.88 \\
\cline{1-14} \cline{2-14}
\bottomrule
\end{tabular}
}
\caption{Detailed performance of 1.3B models on the BLiMP benchmark. These results correspond to the aggregated results reported in Table \ref{tab:tf_main_head}}
\label{tab:blimp_breakdown_1b}
\end{table*}

\begin{table*}[ht]
\resizebox{1\linewidth}{!}{
\begin{tabular}{ccc|ccccccccccc}
\toprule
\toprule
 &  &  & Anaphor & Argument & Binding & Control/ & Ellipsis & Filler & Irregular & Island & NPI & Quantifiers & Subject-Verb \\
Tokens & Heads & Curriculum & Agreement & Structure &  & Raising &  & Gap & Forms & Effects & Licensing &  & Agreement  \\
\midrule
\midrule
\multirow[c]{13}{*}{Subword} & 1 LL & -  & 92.8 & 76.88 & 78.82 & 70.62 & 78.45 & 72.59 & 90.65 & 55.24 & 64.96 & 74.03 & 83.6 \\
\cline{2-14}
 & \multirow[c]{3}{*}{2 LL} & -  & 91.4 & 74.43 & 78.63 & 71.84 & 75.0 & 71.47 & 92.0 & 54.4 & 48.01 & 72.88 & 81.4 \\
 &  & Forward  & 90.65 & 75.57 & 78.62 & 71.4 & 77.15 & 72.4 & 90.5 & 54.3 & 64.56 & 64.05 & 82.2 \\
 &  & Reverse  & 92.55 & 75.58 & 79.48 & 71.78 & 77.15 & 72.21 & 91.5 & 55.97 & 63.79 & 75.72 & 82.92 \\
\cline{2-14}
 & \multirow[c]{3}{*}{2 TL} & -  & 89.7 & 75.56 & 78.85 & 70.06 & 76.3 & 72.04 & 91.6 & 57.3 & 62.37 & 69.1 & 77.73 \\
 &  & Forward  & 89.3 & 74.9 & 78.47 & 71.14 & 75.35 & 71.07 & 85.1 & 50.8 & 65.49 & 73.6 & 80.2 \\
 &  & Reverse  & 89.8 & 76.08 & 80.02 & 72.22 & 76.7 & 72.2 & 88.3 & 57.25 & 65.13 & 66.92 & 80.53 \\
\cline{2-14}
 & \multirow[c]{3}{*}{4 LL} & -  & 92.1 & 72.17 & 78.0 & 69.0 & 73.45 & 70.19 & 90.15 & 52.35 & 70.87 & 68.08 & 73.18 \\
 &  & Forward  & 89.65 & 73.96 & 78.63 & 70.32 & 72.75 & 71.26 & 89.15 & 54.11 & 61.94 & 63.6 & 78.78 \\
 &  & Reverse  & 91.7 & 75.66 & 79.4 & 70.6 & 76.2 & 71.61 & 91.0 & 54.05 & 69.7 & 70.03 & 76.7 \\
\cline{2-14}
 & \multirow[c]{3}{*}{4 TL} & -  & 89.9 & 72.99 & 78.37 & 70.34 & 72.05 & 70.27 & 91.55 & 52.21 & 59.81 & 72.28 & 75.55 \\
 &  & Forward  & 90.75 & 73.42 & 77.95 & 70.12 & 75.2 & 70.73 & 87.15 & 49.51 & 64.71 & 67.8 & 78.45 \\
 &  & Reverse  & 91.65 & 75.0 & 78.28 & 71.24 & 77.2 & 70.36 & 90.65 & 54.76 & 64.81 & 70.78 & 76.88 \\
\midrule
\midrule
\multirow[c]{7}{*}{Byte} & 1 LL & -  & 76.75 & 74.49 & 76.88 & 67.14 & 74.55 & 70.43 & 91.6 & 55.34 & 50.36 & 65.17 & 84.33 \\
\cline{2-14}
 & \multirow[c]{3}{*}{4 LL} & -  & 77.5 & 74.83 & 77.0 & 69.66 & 76.55 & 68.76 & 82.1 & 51.98 & 59.54 & 68.08 & 82.2 \\
 &  & Forward  & 78.8 & 74.84 & 75.98 & 67.98 & 77.25 & 69.26 & 91.05 & 50.78 & 54.03 & 63.05 & 82.7 \\
 &  & Reverse  & 87.7 & 76.14 & 77.47 & 70.44 & 74.7 & 71.01 & 90.25 & 50.94 & 58.07 & 67.92 & 84.5 \\
\cline{2-14}
 & \multirow[c]{3}{*}{8 LL} & -  & 81.6 & 74.7 & 74.08 & 67.98 & 75.2 & 68.27 & 84.25 & 48.49 & 52.16 & 69.85 & 80.83 \\
 &  & Forward  & 78.9 & 74.02 & 75.72 & 67.96 & 76.45 & 68.07 & 87.35 & 52.41 & 56.76 & 62.45 & 79.97 \\
 &  & Reverse  & 79.2 & 75.19 & 75.87 & 70.28 & 77.2 & 71.26 & 85.8 & 51.81 & 60.41 & 71.8 & 81.82 \\
\cline{1-14} \cline{2-14}
\bottomrule
\end{tabular}
}
\caption{Detailed performance of 3B models on the BLiMP benchmark. These results correspond to the aggregated results reported in Table \ref{tab:tf_main_head}}
\label{tab:blimp_breakdown_3b}
\end{table*}

\end{document}